\documentclass[conference]{IEEEtran}
\IEEEoverridecommandlockouts
\usepackage{cite}
\usepackage{amsmath,amssymb,amsfonts}
\usepackage{balance}
\usepackage{graphicx}
\usepackage{textcomp}
\usepackage{xcolor}
\usepackage{booktabs}
\usepackage{floatrow}
\def\BibTeX{{\rm B\kern-.05em{\sc i\kern-.025em b}\kern-.08em
    T\kern-.1667em\lower.7ex\hbox{E}\kern-.125emX}}

\newcommand{\fb}{\mathbf{f}}

\newcommand{\pb}{\mathbf{p}}

\newcommand{\ub}{\mathbf{u}}

\newcommand{\xb}{\mathbf{x}}
\newcommand{\yb}{\mathbf{y}}

\newcommand{\Ab}{\mathbf{A}}

\newcommand{\Cb}{\mathbf{C}}

\newcommand{\Ib}{\mathbf{I}}

\newcommand{\Kb}{\mathbf{K}}

\newcommand{\Qb}{\mathbf{Q}}

\newcommand{\Tb}{\mathbf{T}}

\newcommand{\Wb}{\mathbf{W}}

\newcommand{\Yb}{\mathbf{Y}}

\newcommand{\alphab}{\boldsymbol{\alpha}}

\newcommand{\epsilonb}{\boldsymbol{\epsilon}}

\newcommand{\Ebb}[1]{\mathbb{E}\left[#1\right]}
\newcommand{\Vbb}[1]{\mathbb{V}\left[#1\right]}

\newcommand{\dd}{{(d)}}

\newcommand{\Nt}{N_\ast}

\newcommand{\Kff}{\Kb_{\fb,\fb}}
\newcommand{\Ktf}{\Kb_{\ast,\fb}}
\newcommand{\Kft}{\Kb_{\fb,\ast}}
\newcommand{\Kfu}{\Kb_{\fb,\ub}}
\newcommand{\Kuf}{\Kb_{\ub,\fb}}
\newcommand{\Kuu}{\Kb_{\ub,\ub}}
\newcommand{\Ktu}{\Kb_{\ast,\ub}}
\newcommand{\Kut}{\Kb_{\ub,\ast}}
\newcommand{\Ktt}{\Kb_{\ast,\ast}}

\newcommand{\Wf}{\Wb_{\fb}}
\newcommand{\Wt}{\Wb_{\ast}}

\newcommand{\DWf}{\left(\partial\Wf\right)}

\newcommand{\N}{\mathcal{N}}
\newcommand{\Ocal}{\mathcal{O}}

\newcommand{\R}{\mathbb{R}}

\DeclareMathOperator{\SoR}{SoR}

\DeclareMathOperator{\vecrm}{vec}

\DeclareMathOperator*{\Motimes}{\text{\raisebox{0.0ex}{\scalebox{1.0}{$\bigotimes$}}}}

\usepackage{url}
\usepackage{notoccite}
\usepackage{subcaption}
\usepackage{enumerate}
\usepackage{siunitx}
\usepackage{bm}
\usepackage{algorithm}
\usepackage{algpseudocode}
\usepackage{verbatim}
\usepackage{outlines}
\usepackage{tikz}
\usepackage{pgfplots}
\usepackage{stackengine} 

\usepackage{multirow}
\usepackage{subcaption}

\definecolor{blue}{RGB}{3, 128, 149}
\definecolor{green}{RGB}{59, 151, 52}
\definecolor{red}{RGB}{248, 60, 93}
\definecolor{yellow}{RGB}{247, 179, 51}
\definecolor{lightred}{RGB}{253, 223, 223}
\definecolor{lightblue}{RGB}{222, 243, 253}
\definecolor{lightgreen}{RGB}{191, 216, 213}
\definecolor{lightyellow}{RGB}{246, 234, 197}

\def\addlegendimage{\csname pgfplots@addlegendimage\endcsname}

\begin{document}

\title{Large-scale magnetic field maps
using structured kernel interpolation
for Gaussian process regression
}

\author{\IEEEauthorblockN{Clara Menzen, Marnix Fetter and Manon Kok}    \IEEEauthorblockA{Delft Center for Systems and Control, Delft University of Technology, Delft, The Netherlands \\          Email: c.m.menzen@tudelft.nl, marnixfetter@gmail.com, m.kok-1@tudelft.nl}}

\maketitle

\begin{abstract}
We present a mapping algorithm to compute large-scale magnetic field maps in indoor environments with approximate Gaussian process (GP) regression.
Mapping the spatial variations in the ambient magnetic field can be used for localization algorithms in indoor areas.
To compute such a map, GP regression is a suitable tool because it provides predictions of the magnetic field at new locations along with uncertainty quantification.
Because full GP regression has a complexity that grows cubically with the number of data points, approximations for GPs have been extensively studied.
In this paper, we build on the structured kernel interpolation (SKI) framework, speeding up inference by exploiting efficient Krylov subspace methods.
More specifically, we incorporate SKI with derivatives (D-SKI) into the scalar potential model for magnetic field modeling and compute both predictive mean and covariance with a complexity that is linear in the data points.
In our simulations, we show that our method achieves better accuracy than current state-of-the-art methods on magnetic field maps with a growing mapping area.
In our large-scale experiments, we construct magnetic field maps from up to $\boldsymbol{40 \thinspace 000}$ three-dimensional magnetic field measurements in less than two minutes on a standard laptop.
\end{abstract}


\section{Introduction}
Indoor positioning and navigation in indoor environments is an active and challenging field of research, see e.g.\ \cite{zafari2019survey,yassin2016recent}.
Since the global positioning system (GPS) does not work properly indoors, existing technologies rely on e.g.\ WLAN \cite{he2015wi} or ultra-wideband \cite{alarifi2016ultra}.
In recent years, a novel and promising approach uses the spatial anomalies of the ambient magnetic field that is present indoors, see e.g.\ 
\cite{haverinen2009global,vallivaara2010simultaneous,chung2011indoor,robertson2013simultaneous,berkovich2019coursa,kim2012indoor,torres2015ujiindoorloc}.
Probabilistic algorithms for indoor localization with magnetic field measurements use e.g.\ an extended Kalman filter \cite{viset2022extended} or a particle filter \cite{solin2016terrain} in combination with Gaussian process (GP) regression.
The motivation to use GPs \cite{rasmussen2010gaussian} is the fact that they can be used to construct a magnetic field map from measurements providing a mean and uncertainty information which are both crucial for probabilistic localization algorithms.
However, full GP regression becomes intractable for a large number of data points $N$, so existing approaches for magnetic field mapping have downsampled the data \cite{vallivaara2010simultaneous}, made maps only using data close to a position of interest \cite{nguyen2008local} or have approximated the GP kernel in terms of a number of basis functions \cite{solin2020hilbert}. 
Each of these methods has downsides, which can impact the accuracy of the map:
The first two methods do not use all the data and the latter relies on a sufficient number of basis functions $M_\mathrm{bf}$ to achieve a good approximation of the kernel function \cite{solin2020hilbert}.
\begin{figure}
    \centering
    \includegraphics[width=\columnwidth]{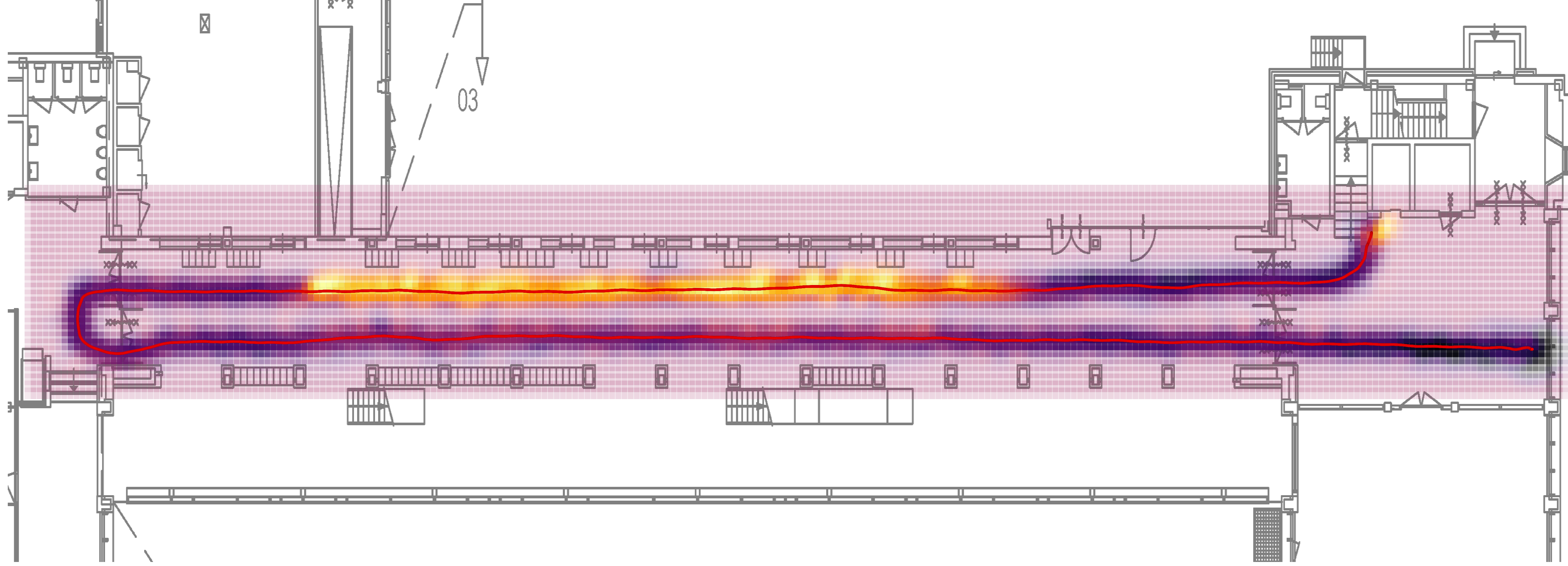}
    \caption{Magnitude of the magnetic field computed by constructing a map from 21\thinspace931 measurements, where darker regions correspond to a higher magnitude. The red line is the walking path, along which measurements are collected.}
    \label{fig:maphallway}
\end{figure}
Inspired by the fact that the literature about approximate GPs offers numerous other approaches for large-scale GPs that overcome the aforementioned limitations, in this paper, we build on the SKI framework by \cite{wilson2015kernel} to construct magnetic field maps.
In the SKI framework, the measurements are observed through $M_\mathrm{ind}$ inducing variables, where the inducing inputs are placed on a Cartesian grid.
The structure of the inducing inputs naturally enables Kronecker structure in the corresponding kernel matrix, as well as structured kernel interpolation (SKI), i.e.\ approximation of kernel matrices by interpolation.
Exploiting Kronecker algebra and the sparsity of the interpolation matrices in Krylov subspace methods, we can compute magnetic field maps in an efficient way.
This allows us to compute large-scale magnetic field maps as illustrated in Fig.\ \ref{fig:maphallway} that are computationally unfeasible for full GP regression on a regular laptop.
To construct the map, we use magnetometer data in combination with positions and orientations that are assumed to be known.
Based on previous work by e.g. \cite{kok2018scalable,solin2018modeling}, we model the magnetic field with the scalar potential model that allows for incorporating physical knowledge into the GP prior.
In this framework, we use SKI with derivatives (D-SKI) \cite{eriksson2018scaling} to compute the predictive means with conjugate gradients. 
For the predictive variance, we adapt the LanczOs Variance Estimates (LOVE) algorithm \cite{pleiss2018constant} to the D-SKI framework. 
The associated computational complexity is \mbox{$\Ocal((J+2T)(3N + M_\mathrm{ind}(M_\mathrm{ind}^{(1)}+M_\mathrm{ind}^{(2)}+M_\mathrm{ind}^{(3)})))$}, where $J$ and $T$ are chosen based on the desired accuracy of the conjugate gradient and Lanczos tridiagonalization algorithm, respectively, and $M_\mathrm{ind}^{(d)}$ is the number of inducing inputs in the $d$th dimension.

\section{Problem formulation}
\label{sec:problem}
We are interested in constructing large-scale magnetic field maps with GP regression.
Similar to \cite{wahlstrom2013modeling,solin2018modeling}, we assume the magnetic field to be curl-free and model the magnetic field measurements as the derivatives of a scalar potential $\varphi$ on which we put a GP prior.
Given $n = 1, \hdots, N$ 3D positions \mbox{$\pb$} at which magnetic field measurements have been collected, the GP model is given by
\begin{equation}
    \label{eq:scalarpotentialmodel}
    \begin{aligned}
    \varphi(\pb) &\sim \mathcal{GP}\left(0, \kappa(\pb,\pb')\right),\\
    \yb_n &= -\nabla \varphi(\pb_n) + \epsilonb_n, \qquad \epsilonb_n \sim \N(\mathbf{0}, \sigma_\yb^2\mathbf{I}_3),
    \end{aligned}
\end{equation}
where $\yb_n\in\mathbb{R}^{3}$ contains the $x$-, $y$- and $z$- component of the magnetic field measurement, $\kappa$ is the kernel function, and $\sigma_\yb^2$ is the noise variance.
We choose the kernel to be the squared exponential kernel, as in related literature, see e.g.\ \cite{vallivaara2011magnetic,vallivaara2010simultaneous,akai2015gaussian,li2015using,lee2020magslam}. 
The squared exponential kernel is given by
\begin{equation}
    \kappa(\pb,\pb^\prime) = \sigma_\fb^2\exp\left(-\frac{\|\pb-\pb^\prime\|_2}{2\ell^2}\right),
\end{equation}
where $\sigma_\fb^2$ and $\ell$ are the hyperparameters of the kernel, the signal variance, and the length scale.

Although we measure the Earth's magnetic field together with anomalies, we choose to only model the anomalies, because this model choice fits better into our approximation scheme.
When also considering the local magnetic field as e.g.\ in \cite{wahlstrom2015modeling}, the kernel includes a linear term as well.

As the gradient operator is a linear operator given the linearity of differentiation~\cite{sarkka2011linear}, the predictive distribution of the three components of the magnetic field in a new location $\pb_*$ can be expressed in terms of a mean and a variance of $\fb_*$ given by
\begin{equation}
\begin{aligned}
    \Ebb{\fb_*} &= \partial^2(\Ktf) \left(\partial^2(\Kff)+\sigma_\yb^2 \Ib_{3N}\right)^{-1}\vecrm\left(\Yb^\top\right),\\
    \Vbb{\fb_*} &= \partial^2(\Ktt) \;- \\
    &\;\;\;\;\;\;\;\partial^2(\Ktf)\left(\partial^2(\Kff)+\sigma_\yb^2 \Ib_{3N}\right)^{-1}\partial^2(\Kft),
    \label{eq:pred}
\end{aligned}    
\end{equation}
where \mbox{$\Yb=[\yb_1^\top\;\yb_2^\top\;\cdots\yb_N^\top]\in\mathbb{R}^{N\times 3}$} are all magnetic field measurements.
The entries of $\partial^2(\Kff)$, $\partial^2(\Ktf)$ and $\partial^2(\Ktt)$ are computed block-wise in terms of $3\times 3$ blocks for each pair of positions with $\nabla_\pb \kappa(\pb,\pb') \nabla_{\pb'}^\top$, $\nabla_\pb \kappa(\pb,\pb_*) \nabla_{\pb_*}^\top$ and $\nabla_{\pb^*} \kappa(\pb_*,\pb_*^\prime) \nabla_{\pb_*^\prime}^\top$, respectively, where $\nabla$ denotes the gradient that is taken w.r.t. to the vector specified in the subscript.

With \eqref{eq:pred} it is possible to predict the magnetic field in new locations.
In practice, however, this is only possible for a small number of data points, since full GP regression generally scales cubically with $N$.
In this case it even scales cubically with $3N$, because of the 3 derivatives.
In this paper, we build on the SKI framework, described in the next section, to make large-scale magnetic field maps in an efficient way.

\section{SKI framework}
\label{sec:SKI}
The SKI framework is based on sparse approximations for GP regression, using a set of $M_\mathrm{ind}$ inducing inputs $\xb_\ub\in\mathbb{R}^{D}$.
In the context of magnetic field modeling, the inducing inputs are positions in $\mathbb{R}^3$.
Based on the Nyström approximation~\cite{williams2000using} of the kernel, the simplest formulation of the inducing input approach is the subset of regressors (SoR), which can be implemented similarly to the predictive distribution for full GP regression using an approximation to the kernel function~\cite{quinonero2005unifying}, given by
\begin{equation}
    \label{eq:kernelmod}
    \kappa_{\SoR}(\xb,\xb^\prime) = \kappa(\xb,\xb_\ub) \Kuu^{-1} \kappa(\xb_\ub,\xb^\prime),
\end{equation}
where $\Kuu$ denotes the covariance matrix of all the inducing inputs.
The approximated kernel function results in new kernel matrices, which are then given by
\begin{subequations}
    \label{eq:modK}
    \begin{equation}
        \Kff = \Kfu \Kuu^{-1} \Kuf,
    \end{equation}
    \begin{equation}
        \Ktt = \Ktu \Kuu^{-1} \Kut,
    \end{equation}
    \begin{equation}
        \Ktf = \Ktu \Kuu^{-1} \Kuf.
    \end{equation}
\end{subequations}

In the SKI framework \cite{wilson2015kernel}, the inducing inputs are placed on a Cartesian 
grid, which is equispaced per dimension and of size $M_\mathrm{ind}^{(1)} \times M_\mathrm{ind}^{(2)} \times \dots \times M_\mathrm{ind}^{(D)}$, for a total of $M_\mathrm{ind} = \prod_{d=1}^D M_\mathrm{ind}^\dd$ inducing inputs.  

Consequently, product kernels - here the squared exponential kernel is considered - decompose over the input dimensions.
Thus, $\Kuu$ can be expressed as a Kronecker product of $D$ matrices ~\cite{wilson2015kernel, saatcci2012scalable}, given by
\begin{equation}
    \Kuu = \Motimes_{d=1}^D \Kuu^\dd,
\end{equation}
where $\Kuu^\dd$ is computed with a squared exponential kernel having a scaled signal variance $\sigma_\fb^{2/D}$ \cite{saatcci2012scalable}.

In addition, in the SKI framework, the cross-covariance matrices $\Kfu$ and $\Ktu$ are approximated using sparse interpolation matrices, $\Wf \in \R^{N \times M}$ and $\Wt \in \R^{\Nt \times M}$, such that 
\begin{equation}
\Kfu \approx \Wf \Kuu \;\;\text{and}\;\; \Ktu \approx \Wt \Kuu.    
\label{eq:interp}
\end{equation}
Each row of the interpolation matrices contains $4^D$ interpolation weights for cubic interpolation \cite{keys1981cubic} which is suggested in \cite{wilson2015kernel}.


\section{Large-scale magnetic field maps}
\label{sec::method}
Our goal is to compute magnetic field maps in 3D using magnetic field measurements as training data.
In order to be able to use large data sets, we exploit mathematical formulations in the SKI framework adapted to magnetic field modeling to compute predictive means and variances in an efficient way.
As the predictive distribution of the scalar potential model in \eqref{eq:pred} is based on the derivatives of the magnetic scalar potential, we approximate the elementwise computation $\partial^2(\cdot)$ of the kernel matrices with D-SKI~\cite{eriksson2018scaling}.
Using D-SKI, $\partial^2(\cdot)$ can be simplified through differentiation of the interpolation scheme, such that it is
\begin{subequations}
\label{eq:dinterpolationscheme}
\begin{align}
    \partial^2(\Kff) \approx \left(\partial\Wf\right)\Kuu\left(\partial\Wf\right)^\top,\\
    \partial^2(\Ktf) \approx \left(\partial\Wt\right)\Kuu\left(\partial\Wf\right)^\top,
\end{align}
\end{subequations}
where $\partial\Wf \in \R^{3N \times M_\mathrm{ind}}$ and $\partial\Wt \in \R^{3\Nt \times M_\mathrm{ind}}$.
Note that the first size of the interpolation matrices is multiplied by a factor of 3 compared to \eqref{eq:interp} due to the three components of the magnetic field.
As mentioned in the previous section, in SKI a cubic interpolation is advised \cite{wilson2015kernel}, while in D-SKI a quintic interpolation scheme is used \cite{eriksson2018scaling}.
We use a cubic interpolation scheme for D-SKI, since preliminary experiments show that the approximation is sufficient for our application.

The predictive distribution in a new location $\pb_*\in\mathbb{R}^{3}$ of the scalar potential model for magnetic field modeling using D-SKI is given by
\begin{subequations}
    \begin{align}
    \Ebb{\fb_*} &=  (\partial \Wt) \Kuu (\partial \Wf)^\top\Ab^{-1}\vecrm\left(\Yb^\top\right)\label{eq:predmean_dski} \\
     \Vbb{\fb_*} &= (\partial\Wt) \; \Kuu \; (\partial\Wt)^\top - (\partial\Wt)\;\Cb\; (\partial\Wt)^\top \label{eq:predvar_dski}
    \end{align}
\end{subequations}
The matrices $\Ab$ and $\Cb$ only depend on the training data and are defined as
\begin{subequations}
    \begin{align}
        \Ab &:=\DWf \Kuu \DWf^\top+\sigma_\yb^2\; \Ib_{3N} \label{eq:defA},\\
         \Cb &:= \Kuu \DWf^\top \Ab^{-1}\DWf \Kuu.\label{eq:defC}
    \end{align}
\end{subequations}

A naive computation of the predictive mean and variance with \eqref{eq:predmean_dski} and \eqref{eq:predvar_dski} would require an inverse of a $3N\times 3N$ matrix.
Inducing inputs on a grid and kernel interpolation, however, enable efficient computations via Krylov subspace methods.
The key to efficient computation is not to construct the matrices involved in \eqref{eq:predmean_dski} and \eqref{eq:predvar_dski} explicitly but to keep them in terms of a factorized format of 3 smaller matrices, one for each input dimension.
Based on \cite{eriksson2018scaling}, we use preconditioned conjugate gradient to find the solution $\alphab$ to the linear system given by
\begin{equation}
\label{eq:linsys_mean}
    \begin{aligned}
    \Ab \; \alphab &= \vecrm(\Yb^\top).
    \end{aligned}
\end{equation}
While in D-SKI fast matrix-vector-multiplications (MVMs) are computed via FFT, we compute them like in \cite{wilson2015kernel} via Kronecker MVMs as described in \cite{wilson2014covariance}.
In this way, the predictive mean of a 3D map is computed with a computational complexity of
\begin{equation}
    \Ocal\left(J\left(3N + M_\mathrm{ind}\sum_{d=1}^3M_\mathrm{ind}^{(d)}\right)\right)=\Ocal_\mathrm{ind},
    \label{eq:complexity}
\end{equation} where $J$ is the number of iterations in the conjugate gradient.
To compute the predictive variance of the magnetic field map, we build on the LanczOs Variance Estimates approach by \cite{pleiss2018constant}, which is based on the Lanczos tridiagonalization algorithm.
We use the LanczOs Variance Estimates within the D-SKI framework to find a low-rank approximation for $\Ab$ given by 
\begin{equation}
    \Ab \approx \Qb_T \Tb_T \Qb_T^\top,
\end{equation}
where $\Qb_T\in\mathbb{R}^{3N\times T}$ contains $T$ orthonormal vectors corresponding to the first $T$ leading eigenvalues and $\Tb_T\in\mathbb{R}^{T\times T}$ has a tridiagonal structure~\cite{golub2013matrix}.
Once an approximation of $\Ab$ is found, $\Cb$ from \eqref{eq:defC} can be computed as
\begin{equation}
\begin{aligned}
    \Cb &= \Kuu (\partial\Wf)^\top \Ab^{-1} (\partial\Wf) \Kuu \\
    &\approx \Kuu (\partial\Wf)^\top \Qb_T \Tb_T^{-1} \Qb_T^\top (\partial\Wf) \Kuu.
    \label{eq:Capprox}
\end{aligned}
\end{equation}
As described in \cite{pleiss2018constant}, we again exploit Kronecker algebra to compute the MVMs in \eqref{eq:Capprox} efficiently.
The computation of $\Cb$ has an associated computational complexity of $\Ocal(2T (3N + M_\textrm{ind} (M_\textrm{ind}^{(1)}+M_\textrm{ind}^{(2)}+M_\textrm{ind}^{(3)})))$ for magnetic field modeling~\cite{pleiss2018constant}.
In numerical implementations, the complexity is higher due to the full reorthogonalization required for the Lanczos tridiagonalization algorithm, scaling linearly with the number of training points $N$ and Lanczos iterations $T$.

Alternatively, the predictive variance could also be computed using conjugate gradient.
However, computing $\Cb$ by solving a linear system to find $\Ab^{-1} (\partial\Wf) \Kuu$ needs to be done for each column of $\left(\partial\Wf\right)\Kuu$ sequentially, which is not particularly efficient.
In~\cite{wilson2015thoughts}, the variance is stochastically estimated by drawing samples from the predictive distributions~\cite{papandreou2011efficient}.
While this approach can reduce the computational complexity associated with the computation of the predictive variances, it introduces significant accuracy losses.

Once $\alphab$ and $\Cb$ are computed, predictions in new locations can be computed with \eqref{eq:predmean_dski} and \eqref{eq:predvar_dski} again by exploiting Kronecker algebra and the structure of the interpolation matrices.
\section{Experiments}
\label{sec:experiments}
In our experiments, we first compare our method to existing ones in simulations with synthetic data, then show our method's scalability with large-scale magnetometer data collected with a motion capture suit.
All computations are done on a 2016 HP ZBook Studio G3 laptop (Intel Core i7 @ 2.60 GHz, 8GB RAM).
\subsection{Accuracy analysis for growing mapping area}
In the first simulation, we compare the accuracy of magnetic field maps computed with our and existing methods.
For this, we create a synthetic data set of 6000 data points, representing a magnetic field map.
Each input is a random 3D vector lying in a box confined by $[-20,\;20]\times [-20,\;20]\times [0.01,\;0.01]$ and the corresponding output is sampled from a GP prior with a curl-free kernel with hyperparameters, length scale, signal variance, and noise variance, $[\ell,\;\sigma_\fb^2,\;\sigma_\yb^2]=[2,\;1,\;0.01]$ which is equivalent to drawing samples from model \eqref{eq:scalarpotentialmodel} \cite{wahlstrom2015modeling}.

\begin{figure}
    \centering
    \input{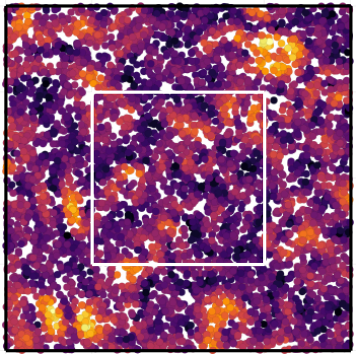}
    \caption{Synthetic data with white and black squares denoting area 1 of size $20\times20$ ($N\approx1526$) and area 2 of size $40\times40$ ($N=6000$). The scattered dots are data points and the color corresponds to the magnitude of the magnetic field.}
    \label{fig:areas}
\end{figure}

We compute two maps with different sizes: area 1 of size $20\times20$ and area 2 of size $40\times40$.
For area 1, a subset of the 6000 data points is used, laying in the white square shown in Fig. \ref{fig:areas}, and for area 2, all data is used.
We divide the data points in each area into 80\% training data and 20\% testing data.
With the testing data, we compute root mean square errors (RMSEs) as a metric for accuracy.

We compare our method to a GP where we downsample the data, as well as to the basis function approach by \cite{solin2020hilbert}, showing how the area size impacts the accuracy of the map for different $M_\mathrm{ind}$, $N_\mathrm{dwn}$ and $M_\mathrm{bf}$ in the respective methods.
Since the domain on which basis functions are computed needs to be a bit bigger than the mapping area \cite{solin2020hilbert}, we add twice the length scale in each dimension.

To be able to compare all methods, we compute the computational complexity of our method $\Ocal_\mathrm{ind}$ as defined in \eqref{eq:complexity},
and impose this complexity as the computational budget for the other methods.
The complexity of the approach of downsampling the data is equal to the cubic complexity of full GP.
The complexity of the basis function approach is linear in the number of data points and quadratic in the number of basis functions.
By equating the complexities 
\begin{equation}
\Ocal_\mathrm{ind}=\Ocal(M_\mathrm{bf}^23N) = \Ocal(N_\mathrm{dwn}^3),
\end{equation}
and solving for $M_\mathrm{bf}$ and $N_\mathrm{dwn}$, those numbers can be used in the corresponding methods.

For each of the two area sizes, we have six different settings, where we vary the number of inducing inputs, i.e., \mbox{$M_\mathrm{ind}^{(1)}=M_\mathrm{ind}^{(2)}=[10,\;20,\;40,\;80,\;100,\;200]$} and $M_\mathrm{ind}^{(3)}=5$.
The number of iterations in the conjugate gradient, $J$, is based on the tolerance for accuracy.
Table \ref{tab:settings} summarizes all settings used in the simulations.
The first two rows are the total number of inducing inputs $M_\mathrm{ind}$ and the other rows are the number of basis functions and downsampled data points that result from the imposed computational budget $\mathcal{O}_\mathrm{ind}$.
We run the simulation 100 times, where each time new data is sampled.
An example of the data is illustrated in Fig.\ \ref{fig:areas}.

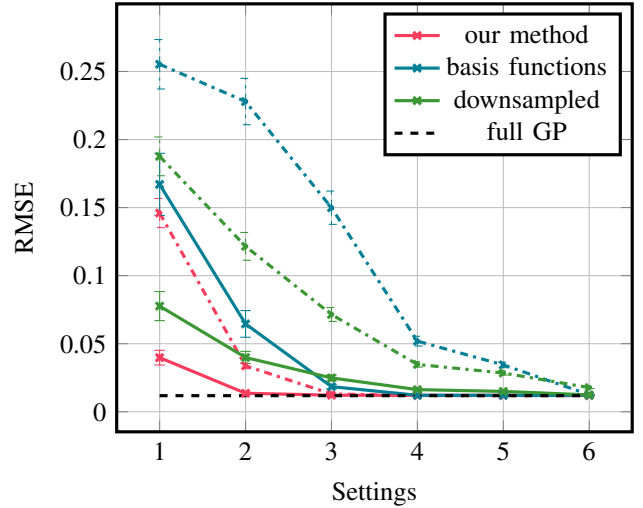
\begin{figure}[h]
    \centering
    \begin{tikzpicture}
\begin{axis}[xmajorgrids, ymajorgrids, xlabel={Settings}, ylabel={RMSE}, xtick={1,2,3,4,5,6}, xticklabels={1,2,3,4,5,6},ytick={0,0.05,0.1,0.15,0.2,0.25}, yticklabels={0,0.05,0.1,0.15,0.2,0.25}, style = {very thick}]
    \addplot[color={red}, mark={x}, error bars/y dir=both, error bars/y explicit]
        coordinates {
            (1,0.039762453031884915)+- (0,0.005402130564340525)
             (2,0.013600418987007045)+- (0,0.0007401000631755177)
             (3,0.012143999602524165)+- (0,0.000600586429570908)
             (4,0.012092895188620747)+- (0,0.0006058913906692767)
             (5,0.01208740695081154)+- (0,0.0006027619904635177)
             (6,0.01208753255698284)+- (0,0.0006025757594871632)
        }
        ;
            \addplot[color={red}, mark={x}, error bars/y dir=both, error bars/y explicit,dashdotted]
        coordinates {
             (1,0.14600586399597337)+- (0, 0.010726956996578335)
             (2,0.03374703046166386)+- (0, 0.002096438186715631)
             (3,0.013085760991173064)+- (0, 0.0003104403029191876)
             (4,0.01188773646438413)+- (0, 0.00027369898596913377)
             (5,0.011849013985593695)+- (0, 0.00027487119343101355)
             (6,0.0118241517594426)+- (0, 0.0002719261778922249)
        }
        ;
    \addplot[color={blue}, mark={x}, error bars/y dir=both, error bars/y explicit]
        coordinates {
             (1,0.1670295684255348)+- (0, 0.02283974567890511)
             (2,0.06454374351412212)+- (0, 0.009794547339327293)
             (3,0.018461296850821308)+- (0, 0.0016918085721066615)
             (4,0.012098782972074425)+- (0, 0.0006101534610281848)
             (5,0.012073040435253413)+- (0, 0.0006016380488559981)
             (6,0.012085562765780608)+- (0, 0.0006018285002911206)
        }
        ;

            \addplot[color={blue}, mark={x}, error bars/y dir=both, error bars/y explicit,dashdotted]
        coordinates {
             (1, 0.25530542863320854)+- (0, 0.01817792407131493)
             (2,0.22790166120109862)+- (0, 0.017041980815931883)
             (3,0.14991696411572822)+- (0, 0.012237861476700099)
             (4,0.05189690084113585)+- (0, 0.0035979848498325193)
             (5,0.03442806894787557)+- (0, 0.002082276752342002)
             (6,0.012525184297789817)+- (0, 0.0003038266547216655)
        }
        ;
    \addplot[color={green}, mark={x}, error bars/y dir=both, error bars/y explicit]
        coordinates {
             (1,0.07762342200907629)+- (0, 0.010678346630716325)
             (2,0.03987107606088282)+- (0, 0.00450135353109769)
             (3,0.024865895075572988)+- (0, 0.0022436817979132697)
             (4,0.016262259305902912)+- (0, 0.0009757606466093278)
             (5,0.014957165874504743)+- (0, 0.0007699096060249318)
             (6,0.012385249450563709)+- (0, 0.0006356722746337382)
        }
        ;
    \addplot[color={green}, mark={x}, error bars/y dir=both, error bars/y explicit,dashdotted]
        coordinates {
             (1,0.1876494309515977)+- (0,  0.014237404451242726)
             (2,0.12152328733699919)+- (0,  0.010224348542517492)
             (3,0.07144771759087927)+- (0,  0.005084212571204043)
             (4,0.03478991313134678)+- (0,  0.0020011076595269467)
             (5,0.02846469496065355)+- (0,  0.0014803049228021025)
             (6,0.017724692430448184)+- (0,  0.0006038035126669136)
        }
        ;
            \addplot[color={black},dashed]
        coordinates {
            (1,0.011823129598306299) 
            (6,0.011823129598306299) 
        }
        ;
    \legend{{our method},{},{basis functions},{},{downsampled},{},{full GP}}
\end{axis}
\end{tikzpicture}
    \caption{RMSE for area 1 (solid line) and 2 (dash-dotted line). For reference, the RMSE of full GP is given for area 2 only, because the other value is very similar. The horizontal axis are the six different settings described in Table \ref{tab:settings}. Mean and standard deviation are plotted for 100 runs of the simulation.}
    \label{fig:rmsevsM}
\end{figure}

Fig.\ \ref{fig:rmsevsM} shows the RMSEs on the testing data computed with the three methods for area 1 (solid line) and area 2 (dash-dotted line).
The mean and standard deviation of the RMSE from 100 runs are plotted.
The horizontal axis denotes the 6 different settings as described in Table \ref{tab:settings}.
The mean RMSE of the full GP is given for area 2 as a reference (dashed black line), for area 1 the RMSE is very similar and therefore not shown.
The figure shows that for each method, the RMSEs are larger for the $40\times40$ area than for the $20\times20$ area.
This implies that a larger $M_\mathrm{ind}$, $N_\mathrm{dwn}$, and $M_\mathrm{bf}$ are required for a larger area to achieve low RMSE.
Also, all methods converge to the RMSE of the full GP in the limit.
Comparing our methods to the other two, for both areas our method has lower RMSEs.
In addition, the difference in RMSEs between our methods and the other methods is more significant for the larger area.
The main takeaway of this simulation is that our method has better accuracy than the other methods when the computational budget $\mathcal{O}_\mathrm{ind}$ is imposed for all methods.
It follows that the computational cost to compute a map of a specified accuracy is lower for our approach compared to the others.
Especially when computing large-scale maps, this becomes important:
Since the number of basis functions needed to approximate the kernel function sufficiently is known to scale with the domain size \cite{solin2020hilbert}, a trade-off between accuracy and computational cost needs to be made.
A similar trade-off is necessary for downsampling the data, because more data points are required for larger areas.

\begin{table*}[t]
    \centering
     \caption{$M_\mathrm{ind}$, $M_\mathrm{bf}$ and $N_\mathrm{dwn}$ used in the 6 settings of the simulation. The first and second rows for basis function approach and downsampled data are the values for the area 1 and 2, respectively. Value ranges for 100 simulations.}
    \begin{tabular}{cccccccc}
    \toprule
         &Setting &1&2&3&4&5&6\\
         \midrule
         \multirow{2}{*}{$M_\mathrm{ind}$}& Area 1 &500 & 2K & 8K & 32K & 50K & 200K\\
         &Area 2&500 & 2K & 8K & 32K & 50K & 200K\\
         \multirow{2}{*}{$M_\mathrm{bf}$} & Area 1 & 35 - 36 & 97 - 98 & 262 - 266 & 726 - 738 & 1012 -  1027 & 2843 - 2887\\
           &Area 2&15 - 16 & 57 - 58 & 152 - 153 & 420 - 422 & 584 - 587 & 1635 - 1646 \\
         \multirow{2}{*}{$N_\mathrm{dwn}$}&Area 1& 55 - 56 & 107 - 108 & 208 - 210 & 411 - 415  & 512 - 518 & 1020 - 1030 \\
                  &Area 2& 49 - 50 & 120 - 121 & 231 - 232 & 455 - 457 & 567 - 569 & 1126 - 1131 \\
         \bottomrule
    \end{tabular}
    \label{tab:settings}
\end{table*}

\begin{figure*}[t]
    \centering
    \input{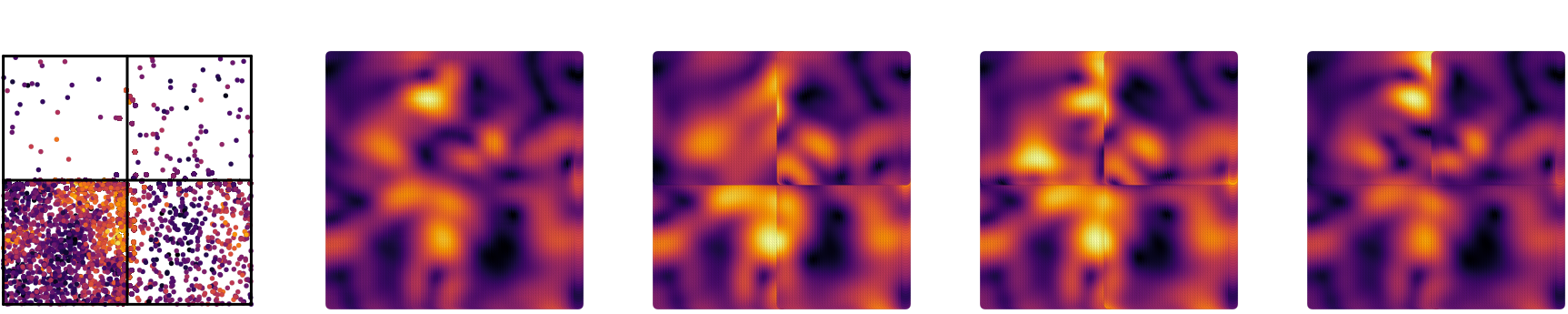}
    \caption{Magnetic field data (a). Predictions with our method for the whole area (b). Predictions with basis functions in smaller areas separately using an overlap of $0$, $0.1\ell$, and $0.3\ell$ for the training data, respectively (c)-(e).}
    \label{fig:fourpred}
\end{figure*}
\subsection{Analysis of maps with divided mapping area}
As mentioned in the previous section, the basis function approach requires a large number of basis functions for growing mapping areas.
To lower the computational complexity, an alternative approach for computing large-scale maps with basis functions is dividing the area into smaller areas for each of which a GP approximation with basis functions is made \cite{kok2018scalable}.
To train each smaller map, not only training data from the mapping area is used, but also training data in close proximity to that area.
In a second simulation, we show that this strategy may result in inconsistencies at boundaries. 
We use synthetic data sampled from a GP prior with a curl-free kernel and with hyperparameters $[\ell,\;\sigma_\fb^2,\;\sigma_\yb^2]=[5,\;1,\;0.01]$, divide the data into four regions and downsample the data by different factors in every region, as shown in Fig.\ \ref{fig:fourpred} (a).
The reason for it is to analyze how the amount of data points in a neighboring area influences the inconsistencies at the boundaries.
For training each map, we use data points in each area as well as data from an overlap of size $0\ell,0.1\ell,0.3\ell$ to the neighboring areas.
The domain on which we compute basis functions is then the size of the mapping area plus the overlap plus twice the length scale.
The result of the reconstruction is shown in Fig.\ \ref{fig:fourpred} (c)-(e).
The figure shows that for no overlap, there are visible inconsistencies in the mean on the magnetic field prediction. 
For $0.1\ell$ and $0.3\ell$, the inconsistencies are smaller but still present.
In addition, the inconsistencies are more visible at the boundaries of areas with fewer data.
As a comparison, a map reconstructed with our method is shown in Fig.\ \ref{fig:fourpred} (b), where the mapping area is not divided.

\subsection{Large-scale map in university building}
\begin{figure*}[t]
    \begin{subfigure}{.25\columnwidth}
    \caption{Magnitude}
        \includegraphics[width=5cm]{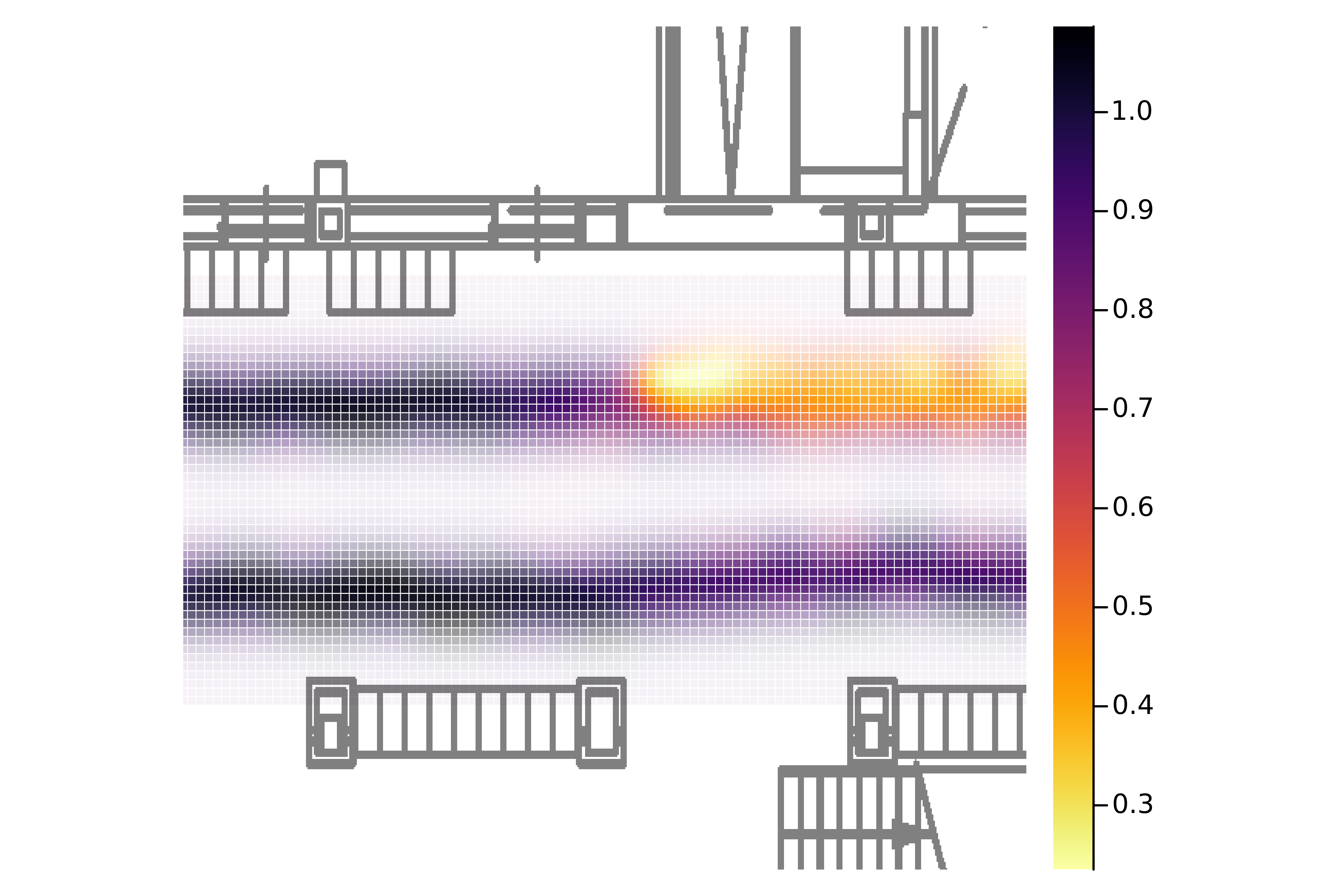}
    \end{subfigure}%
    \begin{subfigure}{.25\columnwidth}
    \caption{$x$-component}
        \includegraphics[width=5cm]{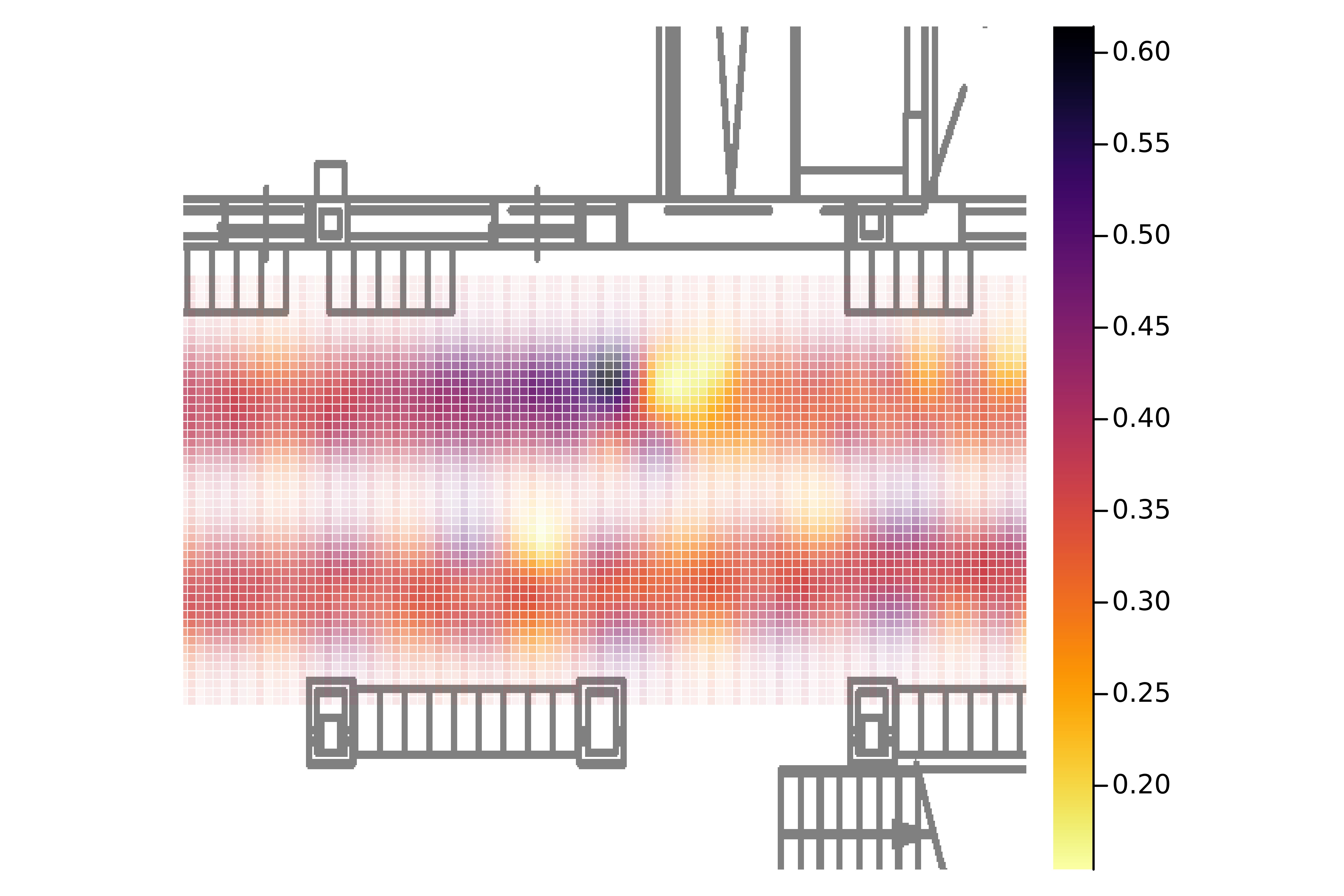}
    \end{subfigure}%
    \begin{subfigure}{.25\columnwidth}
    \caption{$y$-component}
        \includegraphics[width=5cm]{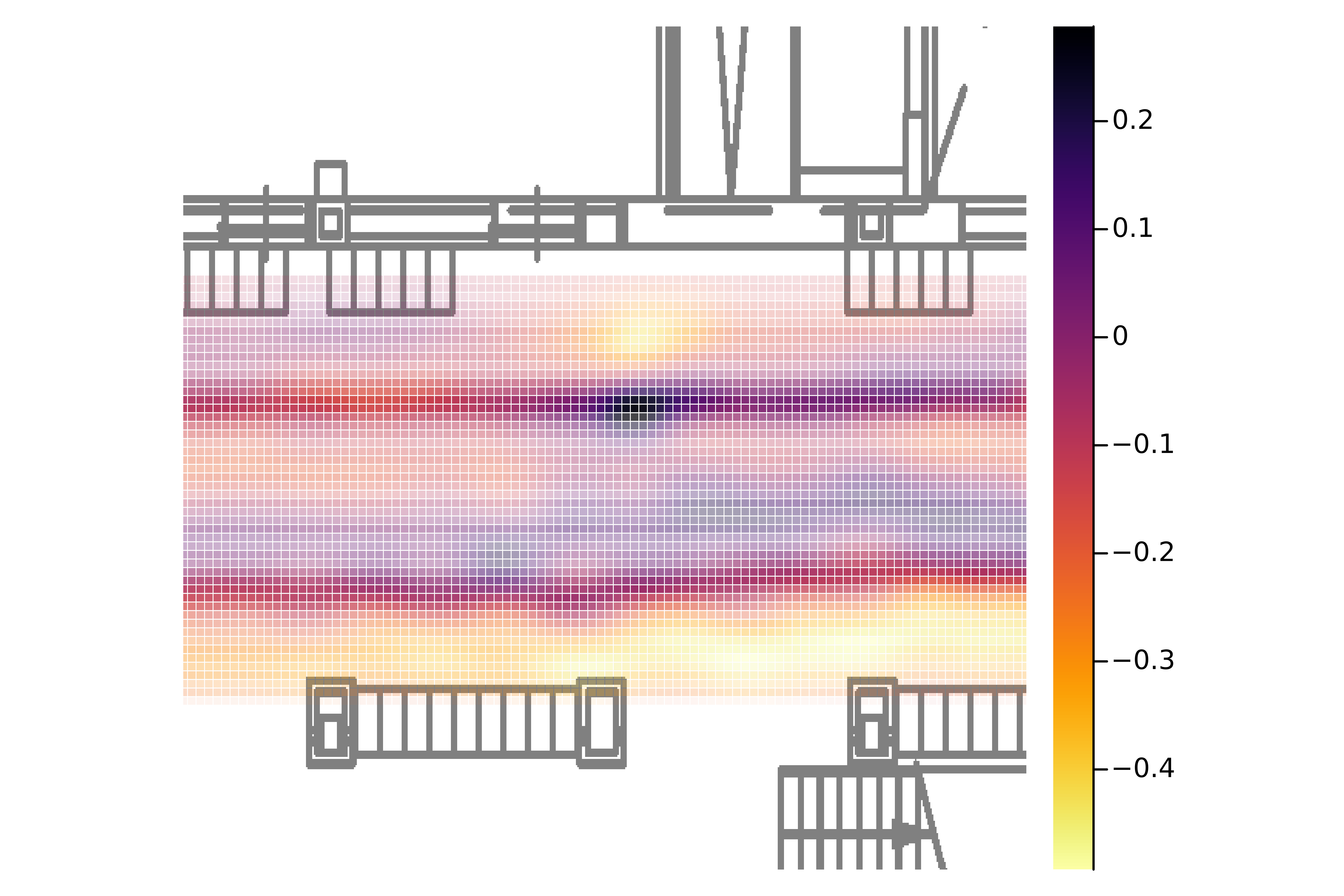}
    \end{subfigure}%
    \begin{subfigure}{.25\columnwidth}
    \caption{$z$-component}
        \includegraphics[width=5cm]{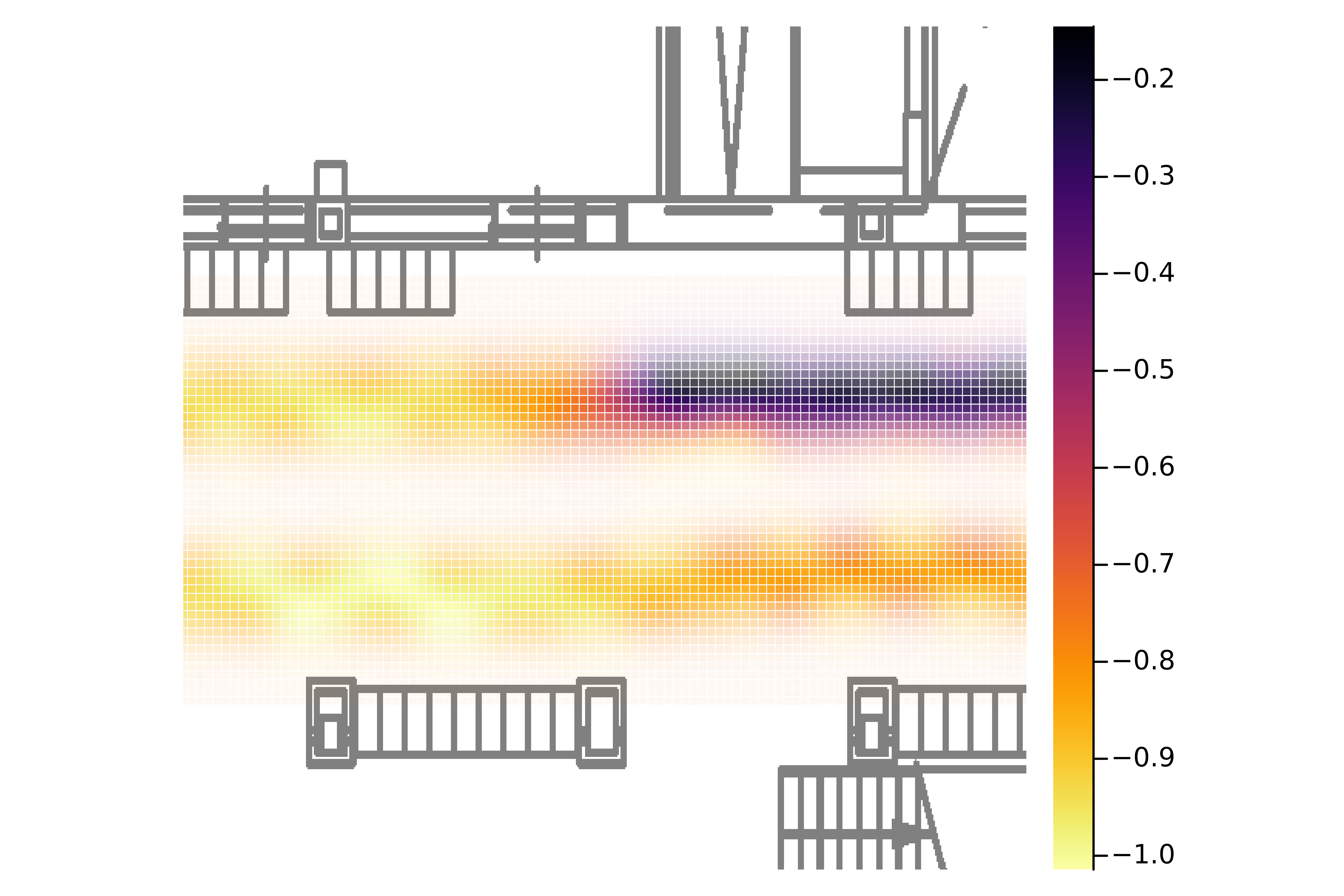}
    \end{subfigure}
    \caption{Magnetic field maps for a smaller region of map in Fig. \ref{fig:maphallway}, located on the left part of the hallway. In the $z$-components, the lockers are strongly visible, indicating that the metal in the lockers mainly disturbs the magnetic field in that direction. Transparency indicates the certainty of the prediction.}
    \label{fig:xyzcomp}
\end{figure*}

For our experiments, we use data collected with a motion capture suit (Xsens MVN Link \cite{XsensSuit}).
The suit contains 17 inertial measurement units (IMUs) equipped with magnetometers tightly attached to segments all over the body. 
The IMUs provide accelerometer, gyroscope, and magnetometer data with position and orientation data in the navigation frame at a maximum sample rate of $\SI{240}{Hz}$.
The magnetometers have been calibrated using software available with the suit, such that after calibration the norm of the undisturbed Earth’s magnetic field is 1 \cite{xsens2022au}.
In a pre-processing step, the data is first rotated to the global frame defined by the magnetic North pole, and second, the mean is subtracted from the $x$-, $y$-, and $z$-component, since we only model the anomalies of the magnetic field.
We use the data collected from one IMU that is located at the pelvis.
In the first large-scale experiment, the magnetic field of one of the university hallway wings at the TU Delft is computed based on $N = 21\thinspace 931$ magnetic field measurements.
The area of interest for this experiment is a rectangle bounded by $[-\SI{34}{m}, \SI{34}{ m}] \times [\SI{-5.25}{m}, \SI{5.25}{m}]$ located at a height of approximately $\SI{1}{m}$.
The magnetic field is estimated with our method using an inducing point grid of size $400 \times 40 \times 4$, for a total of $64\thinspace000$ inducing points. 
The number of inducing points per dimension is chosen such that several inducing points are present per characteristic length scale.
The hyperparameters are trained on a subset of the data by minimizing the log marginal likelihood with the curl-free kernel, resulting in $\ell=\SI{0.5}{m}$, $\sigma_\fb = 0.2$ and $\sigma_\yb=0.01$.
Fig.\ \ref{fig:maphallway} shows the magnitude of the magnetic field predictions, computed from the three components.
Darker regions in the figure correspond to a higher magnitude of the magnetic field. 
Since there are metallic lockers located in the hallway, we expected a strong magnetic anomaly, which is visible in the figure.
Fig. \ref{fig:xyzcomp} shows a smaller section of the magnetic field map, in terms of its magnitude, as well as its $x$-, $y$- and $z$-component.
The magnetic disturbance caused by the lockers is mostly visible in the $z$-components, as shown in the upper right part of Fig.\ \ref{fig:xyzcomp} (d).
The transparency in the figure indicates the certainty of the map.

Regarding computational time, computing the map with $21\thinspace931$ measurements took approximately $\SI{1}{min}$ for training and $\SI{18}{s}$ for testing.
In a second experiment with $41\thinspace383$ data points, the training took approximately $\SI{97}{s}$ for training and $\SI{18}{s}$ for testing.
While with full GP the map would not be feasible to compute on our laptop, our method scales very well: 
The computing time approximately only doubling when doubling the data points, thus is approximately linear in $N$.

\section{Conclusion}
\label{sec:conclusion}
In this paper, we described an algorithm to efficiently compute large-scale magnetic field maps using approximate Gaussian process regression.
We used inducing inputs on a grid and structured kernel interpolation with derivative to compute the predictive mean and an algorithm based on Lanczos tridiagonalization to compute variance estimates.
We compared our method to existing methods in simulations and showed its scalability in large-scale experiments.
There are multiple directions for future work.
The kernel in the scalar potential model can be complemented by a linear kernel as in \cite{wahlstrom2013modeling,solin2018modeling} in order to also model the underlying Earth's magnetic field.
The linear kernel, however, is not a product kernel and can thus not be decomposed as a Kronecker product.
When using the equivalent curl-free model instead, the kernel consists of a constant term and a curl-free kernel \cite{wahlstrom2015modeling}, where the constant term can again be decomposed as a Kronecker product.
In this way, a model considering both the Earth's magnetic field and the spatial anomalies can be used in the D-SKI framework.
In addition, the presented method can be extended to online mapping to enable its use in e.g.\ simultaneous localization and mapping (SLAM) algorithms.
The SKI framework for online GPs has been described in \cite{stanton2021kernel} and can be adapted to magnetic field mapping.

\section*{Acknowledgment}
This publication is part of the project “Sensor Fusion For Indoor Localisation Using The Magnetic Field” with project number 18213 of the research program Veni which is (partly) financed by the Dutch Research Council (NWO).

We would like to thank Fabian Girrbach from Movella Technologies for post-processing data collected with the motion capture suit.

\balance
\bibliographystyle{IEEEtran}
\bibliography{IEEEfull}

\end{document}